\pdfoutput=1
\documentclass[review]{elsarticle}

\usepackage{hyperref}

\usepackage{amsfonts}
\usepackage{amsmath}
\usepackage{algorithm}
\usepackage{algorithmic}
\usepackage{graphicx} 
\usepackage{subfigure}
\usepackage{caption}
\usepackage{multirow}

\journal{Journal of \LaTeX\ Templates}

\begin{document}
	\captionsetup[figure]{labelfont={bf},name={Fig.},labelsep=period}

\begin{frontmatter}

\title{Joint group and residual sparse coding for image compressive sensing
\tnoteref{mytitlenote}}
\tnotetext[mytitlenote]{Fully documented templates are available in the elsarticle package on \href{http://www.ctan.org/tex-archive/macros/latex/contrib/elsarticle}{CTAN}.}

\author[mymainaddress]{Lizhao Li}
\author[mymainaddress]{Song Xiao}

\address[mymainaddress]{State Key Lab of Integrated Services Networks, Xidian University, Xi'an 710071, China}

\begin{abstract}
Nonlocal self-similarity and group sparsity have been widely utilized in image compressive sensing (CS). However, when the sampling rate is low, the internal prior information of degraded images may be not enough for accurate restoration, resulting in loss of image edges and details. In this paper, we propose a joint group and residual sparse coding method for CS image recovery (JGRSC-CS). In the proposed JGRSC-CS, patch group is treated as the basic unit of sparse coding and two dictionaries (namely internal and external dictionaries) are applied to exploit the sparse representation of each group simultaneously. The internal self-adaptive dictionary is used to remove artifacts, and an external Gaussian Mixture Model (GMM) dictionary, learned from clean training images, is used to enhance details and texture. To make the proposed method effective and robust, the split Bregman method is adopted to reconstruct the whole image. Experimental results manifest the proposed JGRSC-CS algorithm outperforms existing state-of-the-art methods in both peak signal to noise ratio (PSNR) and visual quality. 
\end{abstract}


\end{frontmatter}

\section{Introduction}
Compressive sensing \cite{D.L.Donoho2006,E.J.Candes,Qaisar2013}-also known as compressed sensing- is a novel framework for signal processing and compression. It states that if a signal is sparse in some domains, we can perfectly recover it from fewer samples or measurements than Nyquist rate. This indicates that we are able to sample and compress signal at the same time. Due to its advantages of down-sampling and accurate recovery, compressive sensing has been widely applied in many fields, such as digital imaging \cite{Singlepixelcam}, channel estimation \cite{Bajwa2010}, wireless sensor network \cite{Luo2009}, medical imaging \cite{Lustig2008} and remote sensing \cite{Alonso2010}.

Suppose a finite length signal $ x \in {\mathbb{R}^n}$ and its measurement $ y \in {\mathbb{R}^m}$ generated by linear projection:
\begin{equation}
y = \Phi x,
\end{equation}
where $\Phi  \in {\mathbb{R}^{m \times n}}(m \ll n)$ is a random sensing matrix. Since $m \ll n$, recovering $ x $ from $ y $ is an ill-posed problem. However, if $ x $ can be sparsely represented in some basis $\Psi  \in {\mathbb{R}^{n \times n}}$ and the sensing matrix $ \Phi $ meets the restricted isometry property (RIP) \cite{D.L.Donoho2006,Qaisar2013}, we can reconstruct the original signal by solving this optimization problem:
\begin{equation}
\arg \mathop {\min }\limits_\theta {\left\| \theta  \right\|_0}{\rm{    }}\quad s.t.{\rm{  }}\quad y = \Phi \Psi \alpha,
\end{equation}
where ${\left\|  \cdot  \right\|_0}$ is a pseudo norm, counting the non-zero entries of its argument. 

However, since ${\left\|  \cdot  \right\|_0}$ is non-convex, solving Eq.(2) is an NP-hard problem. So the ${l_0}$ norm is often replaced by the ${l_1}$ norm:
\begin{equation}
\arg \mathop {\min }\limits_\theta {\left\| \theta  \right\|_1}{\rm{    }}\quad s.t.{\rm{  }}\quad y = \Phi \Psi \alpha.
\end{equation}

Eq.(3) can be transformed to Lagrangian form:
\begin{equation}
\alpha {\rm{ = }}\mathop {\arg \min }\limits_\alpha  \left\| {y - \Phi \Psi \alpha } \right\|_2^2 + \lambda {\left\| \alpha  \right\|_1},
\end{equation}
where $ \left\| {y - \Phi \Psi \alpha } \right\|_2^2 $ is the cost function and $ \lambda $ denotes the regularization parameter. Eq.(4) can be solved by various algorithms, such as split Bregman algorithm \cite{SBI} and alternative direction multiplier method (ADMM).

Since most natural images have priori characteristics, the optimization problem of image compressive sensing can be formulated as:
\begin{equation}
\mathop {\arg \min }\limits_u \left\| {y - \Phi u} \right\|_2^2 + \lambda \mathcal{R}(u),
\end{equation}
where $ \mathcal{R}(u) $ is the regularization item that represents the prior information of images. Conventional image prior, such as total variation (TV) that characterize the local smoothness of images, has been employed for image CS \cite{TVAL3}. But it may favor piecewise constant solution, resulting in over-smooth. To overcome this problem, many methods have been proposed. For example, Candes \textit{et al.} \cite{reweightedl1}presented the weighted total variation to enhance the sparsity of TV norm. In \cite{ImprovedTV}, Zhang \textit{et al.} proposed a framework that introduced nonlocal means (NLM) into traditional TV. Chen \textit{et al.} \cite{frtv} combined fractional-order total variation with image sparsity regularization, and obtained better PSNR than \cite{TVAL3}.

Recently, patch-based nonlocal similarity has shown its potential in image processing\cite{BM3D}\cite{BM3DCS}\cite{Mairal2009}\cite{CoS}\cite{JASRCS}. As an extension of the BM3D (Block-Matching and 3D filtering) denoising algorithm\cite{BM3D}, BM3D-CS \cite{BM3DCS} introduced 3D collaborative filter into the CS framework, and brought obvious improvement to the recovery quality. Eslahi \textit{et al.} \cite{JASRCS} combined 3D sparsity filter with local sparsity, proposing a new regularization called joint adaptive sparsity regularization (JASR). Elad \textit{et al.} \cite{Elad2006} proposed a patch-based sparse representation algorithm for image denoising, leading to state-of-the-art denoising performance. Motivated by \cite{Elad2006}, many patch-based sparse coding methods for image CS have been proposed \cite{Dong20121109} \cite{NCSR} \cite{Eslahi201688} \cite{ALSB}. For instance, Dong \textit{et al.} \cite{Dong20121109} combined patch sparsity estimation with weighted nonlocal self-similarity constraint to balance the adaptation and robustness of the proposed algorithm. In \cite{ALSB},  the sparsity of natural images is characterized by non-convex patch-based sparse coding, and a new framework is proposed to solve the ${\rm{L}}0$ minimization problem.

More Recently, instead of image patch, patch group is used as the basic unit of sparse coding, and achieves better performance than patch-based algorithms \cite{NIPS2010_3997} \cite{SGSR} \cite{GSR} \cite{GSR-NCR}. In \cite{SGSR}, structural group sparsity representation (SGSR) is proposed to characterize both local and nonlocal similarity of images. Zha \textit{et al.} \cite{GSR-NCR} incorporate a non-convex penalty function to group sparse representation, and obtain state-of-the-art reconstruction performance.

However, most previous methods for CS image reconstruction only consider internal prior information. In this paper, we incorporate external and internal prior into a unified framework, and propose a joint group and residual sparse coding method for CS image reconstruction (JGRSC-CS). In the proposed JGRSC, a patch group and its residual are encoded with internal and external dictionaries respectively. For each group, the internal dictionary is generated by singular value decomposition (SVD), and the external dictionary is learned from clean images based on Gaussian Mixture Model (GMM). To make the algorithm tractable, the split Bergman method is employed to efficiently solve the optimization problem. Experimental results show that the proposed algorithm outperforms many state-of-the-art algorithms in terms of PSNR  and visual perception quality.

The rest of this paper is organized as follows. Section 2 presents a brief introduction of group sparse coding and Gaussian mixture model. In Section 3, we elaborate the joint group and residual sparse coding method for CS image recovery. Experimental results are presented in Section 4. In Section 5, we conclude the paper.

\section{Background}
\subsection{Group sparse coding}
Patch-based sparse coding assumes that every image patch could be sparsely represented by an over-completed dictionary. Suppose an image $ x \in {\mathbb{R}^N}$ and a patch ${x_i}$ of size $\sqrt n  \times \sqrt n $ at location i, $i = 1,2, \ldots , N$. Noting that all patches are overlapped. Then we have
\begin{equation}
{x_i} = {R_i}(x).
\end{equation}
${R_i}(\cdot)$ is an operator extracting the $i$th patch from the image. For every patch, given a dictionary ${D_i}$, it can be written as 
\begin{equation}
{x_i} = {D_i}{\alpha _i}.
\end{equation}

So the whole image can be reconstructed from 
\begin{equation}
x \approx {(\sum\limits_i^N {R_i^T{R_i}} )^{ - 1}}(\sum\limits_i^N {R_i^T{D_i}{\alpha _i}} ).
\end{equation}

Patch-based sparse coding methods ignore the relationship between similar patches. To overcome this 
disadvantage, group sparse coding is proposed. Instead of single patch, group sparse coding treats the patch group as the basic unit of sparse coding. For each patch ${x_i}$, we search its $(m - 1)$ most similar patches within a searching window, and stack these patches into a matrix ${x_{{G_i}}} \in {R^{n \times m}}$. Every group is encoded with a dictionary ${D_{{G_i}}}$, then we can recovery the image by averaging all the patches
\begin{equation}
x \approx {(\sum\limits_i^N {R_{{G_i}}^T{R_{{G_i}}}} )^{ - 1}}(\sum\limits_i^N {R_{{G_i}}^T} {D_{{G_i}}}{\alpha _{{G_i}}}),
\end{equation}
where ${R_{{G_i}}}$ is the the matrix that extracts the most matched patches of ${x_i}$, and ${\alpha _{{G_i}}}$ is the sparse coefficient of patch group ${x_{{G_i}}}$.

\subsection{Gaussian mixture model}
For a single variable $x$ that follows the Gaussian distribution, it can be modeled as
\begin{equation}
\mathcal N(x\left| {\mu ,{\sigma ^2}} \right.) = \frac{1}{{(2\pi {\sigma ^2})}}{e^{ - \frac{1}{{2{\sigma ^2}}}{{(x - \mu )}^2}}},
\end{equation}
where $\mu $ is the mean and ${\sigma ^2}$ is the variance. In the case of a vector ${\rm{x}} \in {{\rm{R}}^N}$, its Gaussian distribution takes the form
\begin{equation}
\mathcal N({\rm{x}}\left| {{\rm{\mu }},\Sigma } \right.) = \frac{1}{{{{(2\pi )}^{\frac{N}{2}}}{{\left| \Sigma  \right|}^{\frac{1}{2}}}}}{e^{ - \frac{1}{2}{{({\rm{x - \mu }})}^{\rm{T}}}{\Sigma ^{ - 1}}({\rm{x - \mu }})}},
\end{equation}
where ${\rm{\mu }}$ is a N-dimensional mean vector and $\Sigma$ is a covariance matrix of size $N \times N$. Gaussian mixture model (GMM) is a linear combination of Gaussian distributions
\begin{equation}
Pr({\rm{x}}) = \sum\limits_{k = 1}^K {{\pi _k}\mathcal N({\rm{x}}\left| {{\mu _k},{\Sigma _k}} \right.)}.
\end{equation}

$Pr({\rm{x}})$ is a superposition of $K$ Gaussian components. In Eq.(12), $\mathcal N({\rm{x}}{\left| {{\mu _k},\Sigma } \right._k})$ is a component of the mixture. ${\mu _k}$ and ${\Sigma _k}$ are the mean and covariance of the $i{\rm{th}}$ component, respectively. ${\pi _k}$ are the normalized mixing coefficients
\begin{equation}
\sum\limits_{k = 1}^K {{\pi _k}}  = 1.
\end{equation}
Since GMM has been successfully used in various inverse problems \cite{Zoran2011GMM} \cite{Yu2012} \cite{yang2014compressive} \cite{yang2014video} \cite{GMMImage} \cite{PGD}  \cite{xu2018}, we will adopt it to train the external dictionary.

\section{The proposed method}
Most existing image compressive sensing methods only consider the nonlocal similarity of the processed image itself, and few utilizes the nonlocal prior of external clean images. In this section, we propose a joint group and residual sparse coding method for CS image recovery, and an efficient framework is developed to solve the optimization problem.

\subsection{Training external dictionary by GMM}
The external dictionary for residual sparse coding is trained from clean images. As mentioned in Section 2.1, for a image patch, we find its $(M - 1)$ most matched patches to form a group ${{\rm{x}}_m}$. Then we subtract the mean ${\mu _m}$ of this group and get the residual group
\begin{equation}
{\overline {\rm{x}} _m} = {{\rm{x}}_m} - {\mu _m},m = 1 \ldots M.
\end{equation}

We collect N residual groups from clean images
\begin{equation}
{{\rm{\overline X}}_n} = {\overline x_{m,n}},n = 1 \ldots N.
\end{equation} 

Considering that GMM has been widely used in image processing, we apply the method mentioned in \cite{Zoran2011GMM}\cite{PGD}\cite{xu2018} to learn the prior, and our goal is learning $K$ Gaussian components from these $N$ groups. Supposing that patches in ${\overline {\rm{X}} _n}$ follows the same Gaussian component, the likelihood of $\{ {\overline {\rm{X}} _n}\} $ is
\begin{equation}
Pr ({\overline X_n}) = \sum\limits_{k = 1}^K {{\pi _k}\prod\limits_{m = 1}^M {N({{\overline x}_{m,n}}\left| {{\mu _k}} \right.,{\Sigma _k})} }. 
\end{equation}

Assuming that all the residual groups are independent, the likelihood function is
\begin{equation}
L = \prod\limits_{n = 1}^N {\Pr ({{\overline X}_n})} .
\end{equation}

According to the Maximum Likelihood Estimation (MLE), we maximize the log function of Eq.(15):
\begin{equation}
\ln L = \sum\limits_{n = 1}^N {\ln \Pr ({{\overline X }_n})}. 
\end{equation}

After initializing the means ${{\rm{\mu }}_k}$, covariances ${\Sigma _k}$, mixing coefficients ${{\rm{\pi }}_k}$ and the value of the log likelihood, Eq.(16) can be optimized using the expectation-maximization (EM) algorithm. In the E step, we then calculate the posterior probability with current parameter values
\begin{equation}
{\gamma _{n,k}} = \frac{{{\pi _k}\prod\limits_{m = 1}^M {\mathcal N({{\overline {\rm{x}} }_{n,m}}{{\left| {\rm{\mu }} \right.}_k},{\Sigma _k})} }}{{\sum\limits_{l = 1}^K {{\pi _l}\prod\limits_{m = 1}^M {\mathcal N({{\overline {\rm{x}} }_{n,m}}{{\left| {\rm{\mu }} \right.}_l},{\Sigma _l})} } }},
\end{equation}
where ${\gamma _{n,k}}$ is the probability that ${\overline {\rm{X}} _n}$ belongs to the $k{\rm{th}}$ component. In the M step, we estimate ${\mu _k}$, ${\Sigma _k}$ and ${\pi _k}$ using ${\gamma _{n,k}}$
\begin{equation}
{N_k} = \sum\limits_{n = 1}^N {{\gamma _{n,k}}}, 
\end{equation}

\begin{equation}
{\mu _k^{new}} = \frac{1}{{{N_k}}}\sum\limits_{n = 1}^N {{\gamma _{n,k}}} \sum\limits_{m = 1}^M {{{\overline {\rm{x}} }_{n,m}}} ,
\end{equation}

\begin{equation}
{\Sigma _k^{new}} = \frac{1}{{{N_k}}}\sum\limits_{n = 1}^N {{\gamma _{n,k}}\sum\limits_{m = 1}^M {{{\overline {\rm{x}} }_{n,m}}\overline {\rm{x}} _{n,m}^T} } ,
\end{equation}

\begin{equation}
{\pi _k^{new}} = \frac{{{N_k}}}{N}.
\end{equation}

We alternate these two steps until the result of Eq.(18) converges, and then the $K$ Gaussian components will be obtained.

\subsection{Joint group and residual sparse coding model}
For a patch group ${x_G} \in {R^{n \times m}}$, the sparse coding model over a given dictionary ${D_G}$ can be formulated as
\begin{equation}
\mathop {\arg \min }\limits_{{\alpha _G}} \frac{1}{2}\left\| {{x_G} - {D_G}{\alpha _G}} \right\|_2^2 + \lambda {\left\| {{\alpha _G}} \right\|_0}.
\end{equation}
where $\alpha $ is the sparse coefficient and $\lambda $ is the regularization parameter. In a patch group, the first column ${x_{{G_1}}}$ is the reference patch and ${x_{{G_i}}},i = 2, \ldots m$ are the $m - 1$ most matched patches of ${x_{{G_1}}}$. To obtain a better estimation of ${x_G}$, in the proposed model, we divide the patch group into two parts
\begin{equation}
{x_G} = \overline {{x_G}}  + {x_{Gr}},
\end{equation}
where $\overline {{x_G}} $ is the mean of all patches and can be calculated as
\begin{equation}
\overline {{x_G}}  = \frac{1}{m}\sum\limits_{i = 1}^m {{x_{{G_i}}}},
\end{equation}
and ${x_{Gr}}$ represents the residual of the group. After selecting a proper dictionary, ${x_{Gr}}$ can be reconstructed via
\begin{equation}
\mathop {\arg \min }\limits_{{\alpha _{Gr}}} \frac{1}{2}\left\| {{x_{Gr}} - {D_{Gr}}{\alpha _{Gr}}} \right\|_2^2 + \lambda {\left\| {{\alpha _{Gr}}} \right\|_1}.
\end{equation}

Incorporating Eq.(27) into Eq.(24), we can obtain the proposed model
\begin{equation}
\mathop {\arg \min }\limits_{{\alpha _G},{\alpha _{Gr}}} \frac{1}{2}\left\| {(\overline {{x_G}}  + {x_{Gr}}) - {D_G}{\alpha _G}} \right\|_2^2 + \frac{1}{2}\left\| {{x_{Gr}} - {D_{Gr}}{\alpha _{Gr}}} \right\| + {\lambda _1}{\left\| {{\alpha _G}} \right\|_0} + {\lambda _2}{\left\| {{\alpha _{Gr}}} \right\|_1}.
\end{equation}

We propose a simple alternating method to solve Eq.(28). For a patch group, we calculate its mean via Eq.(26) and obtain its residual matrix by subtracting it from the original matrix. For fixed ${\alpha _G}$, the ${\alpha _{Gr}}$-subproblem is
\begin{equation}
\mathop {\arg \min }\limits_{{\alpha _{Gr}}} \frac{1}{2}\left\| {{x_{Gr}} - {D_{Gr}}{\alpha _{Gr}}} \right\|_2^2 + {\lambda _2}{\left\| {{\alpha _{Gr}}} \right\|_1}.
\end{equation}

Similar to \cite{Zoran2011GMM} \cite{Yu2012} \cite{PGD}, we assume that ${x_{Gr}}$ follows the Gaussian distribution and select its most matched Gaussian from the trained mixture. The probability of every component can be 
calculated as
\begin{equation}
\Pr (k\left| {{x_{Gr}}} \right.) = \frac{{\prod\limits_{i = 1}^m {N(x_{Gr}^i\left| {0,{\Sigma _k} + \sigma _n^2} \right.I)} }}{{\sum\limits_{j = 1}^K {\prod\limits_{i = 1}^m {N(x_{Gr}^i\left| {0,{\Sigma _j} + \sigma _n^2} \right.I)} } }},
\end{equation}
where ${\sigma _n}$ is the variance Gaussain white noise.The component with the most highest probability will be selected to generate the dictionary, and the dictionary can be produced by singular value decomposition
\begin{equation}
\Sigma_{k}  = {D_{Gr}}{\Lambda _{Gr}}D_{Gr}^T,
\end{equation}
where ${D_{Gr}}$ is the an orthonormal matrix composed of the eigenvectors and ${\Lambda _{Gr}}$ is the diagonal matrix of eigenvalues. Since ${D_{Gr}}$ can represent the structural variations of the selected component, we use it as the dictionary of residual sparse coding.

Now we go back to Eq.(29). Under the framework of Bayesian, the MAP of ${\alpha _{Gr}}$ with ${x_{Gr}}$ is
\begin{align}
{\alpha _{Gr}} &= \mathop {\arg \max }\limits_{{\alpha _{Gr}}} \log P({\alpha _{Gr}}\left| {{x_{Gr}}} \right.) \nonumber\\
&=\mathop {\arg \max }\limits_{{\alpha _{Gr}}} \{ \log P({x_{Gr}}\left| {{\alpha _{Gr}}} \right.) + \log P({\alpha _{Gr}})\}.
\end{align}

Assuming ${x_{Gr}}$ is characterized by the Gaussian noise of ${\sigma _n}$ and the sparse coefficient ${\alpha _{Gr}}$ follows i.i.d Laplacian distribution, we obtain
\begin{equation}
\mathop {\arg \min }\limits_{{\alpha _{Gr}}} \frac{1}{2}\left\| {{x_{Gr}} - {D_{Gr}}{\alpha _{Gr}}} \right\|_2^2 + 2\sqrt 2\sigma _n^2 \times \sum\limits_{i = 1}^n {\frac{1}{{{\sigma _i}}}\left| {\alpha _{Gr}^i} \right|},
\end{equation}
where ${\sigma _i}$ is the standard deviations of $\alpha _{Gr}^i$. By comparing Eq.(29) with Eq.(33), we can see that ${\lambda _2} = \frac{{2\sqrt 2 \sigma _n^2}}{{{\sigma _i}}}$. So Eq. (29) admits a close-form solution
\begin{equation}
{\alpha _{Gr}} = {\mathop{\rm sgn}} (D_{Gr}^T{x_{Gr}}) \bullet \max (\left| {D_{Gr}^T{x_{Gr}}} \right| - \frac{{2\sqrt 2 {\lambda _2}\sigma _n^2}}{{{\sigma _i}}},0).
\end{equation}

For fixed ${\alpha _{Gr}}$, the ${\alpha _G}$-subproblem is
\begin{equation}
\mathop {\arg \min }\limits_{{\alpha _G}} \frac{1}{2}\left\| {(\overline {{x_G}}  + {x_{Gr}}) - {D_G}{\alpha _G}} \right\|_2^2 + {\lambda _1}{\left\| {{\alpha _G}} \right\|_0}.
\end{equation}

Applying the singular value decomposition (SVD) to $\left( {\overline {{x_G}}  + {x_{Gr}}} \right)$, we have
\begin{equation}
\left( {\overline {{x_G}}  + {x_{Gr}}} \right) = {U_G}{\Sigma _G}V_G^T,
\end{equation}
where ${\Sigma _G}$ is a diagonal matrix formed by the eigenvalues. The adaptive internal dictionary is defined as 
\begin{equation}
{D_G}{\rm{ = }}{U_G}V_G^T.
\end{equation}

So Eq.(33) has a close-form solution 
\begin{equation}
\alpha _G^{\rm{i}} = hard(\Sigma _G^i,\sqrt {2{\lambda _1}} ) = \Sigma _G^i\bullet
(\left| {\Sigma _G^i} \right| - \sqrt {2{\lambda _1}} ),
\end{equation}
where $hard$ is hard thresholding function \cite{Hard} and $\bullet$ represents the element-wise product. After getting ${\alpha _G}$, we can reconstruct the group by $\widetilde {{x_G}} = {D_G}{\alpha _G}$.

\subsection{CS reconstruction via joint group and residual sparse coding}
In this section, we rewrite Eq.(2) as 
\begin{equation}
\mathop {\arg \min }\limits_x \frac{1}{2} \left\| {y - \Phi x} \right\|_2^2 + \lambda {\left\| \alpha  \right\|_0}{\rm{  }}\quad s.t.{\rm{  }}\quad x = D\alpha.
\end{equation}
Its unconstrained form is
\begin{equation}
\mathop {\arg \min }\limits_x \frac{1}{2} \left\| {y - \Phi x} \right\|_2^2 + \lambda {\left\| \alpha  \right\|_0} + \frac{\mu }{2}\left\| {x - D\alpha }. \right\|_2^2
\end{equation}

Eq.(40) can be effectively solved by the split Bergman iteration (SBI) method \cite{SBI}. The main idea
of the SBI is to split an unconstrained problem to several subproblems and Bergman iteration. Applying the
SBI framework to Eq.(40), it is converted to the following three iterations:
\begin{equation}
{x^{(l + 1)}} = \mathop {\arg \min }\limits_x \frac{1}{2}\left\| {y - \Phi x} \right\|_2^2 + \frac{\mu }{2}\left\| {x - D{\alpha ^{(l)}} - {b^{(l)}}} \right\|_2^2,
\end{equation}
\begin{equation}
{\alpha ^{(l + 1)}} = \mathop {\arg \min }\limits_x \lambda {\left\| \alpha  \right\|_0} + \frac{\mu }{2}\left\| {{x^{(l + 1)}} - D\alpha  - {b^{(l)}}} \right\|_2^2,
\end{equation}
\begin{equation}
{b^{(l + 1)}} = {b^{(l)}} - ({x^{(l + 1)}} - D{\alpha ^{(l + 1)}}),
\end{equation}
where $b$ is an auxiliary variable and $l$ is the iteration number. Eq.(40) is transformed to $x$ subproblem and $\alpha$ subproblem. In the following, we will show how to solve
these subproblems efficiently. To avoid confusion, the superscript $l$ will be omitted.

\subsubsection{$x$ subproblem}
For a fixed $\alpha$, the $x$ subproblem becomes:
\begin{equation}
{x} = \mathop {\arg \min }\limits_x \frac{1}{2}\left\| {y - \Phi x} \right\|_2^2 + \frac{\mu }{2}\left\| {x - D{\alpha } - {b}} \right\|_2^2,
\end{equation}

Eq.(44) is a quadratic optimization problem and its close-form solution is
\begin{equation}
x = {({\Phi ^{\rm{T}}}\Phi  + \mu I)^{ - 1}}({\Phi ^{\rm{T}}}y + \mu D\alpha  + \mu b),
\end{equation}
where $I$ is identity matrix. However, $\Phi$ is a random matrix, and it is costly to invert $({\Phi ^{\rm{T}}}\Phi  + \mu I)$. In practice, it can be accelerated by utilizing the gradient descent method:
\begin{equation}
x = x - \eta  \cdot \nabla,
\end{equation}
where $\eta$ is the step size and $\nabla$ represents the gradient direction of Eq.(44). Therefore, we can update $x$ by calculating:
\begin{equation}
x = x - \eta ({\Phi ^{\rm{T}}}\Phi x - {\Phi ^{\rm{T}}}y + \mu x - \mu D\alpha  - \mu b).
\end{equation}

\subsubsection{$\alpha$-subproblem}
For a fixed $x$, the $\alpha$ subproblem is
\begin{equation}
\alpha  = \mathop {\arg \min }\limits_\alpha  \frac{1}{2}\left\| {x - D\alpha  - b} \right\|_2^2 + \frac{\lambda }{\mu }{\left\| \alpha  \right\|_0}.
\end{equation}

We define ${x_n} = x - b$, and ${x_n}$ can be seen as the noisy observation of $x$. So Eq.(48) can be rewritten as
\begin{equation}
\alpha  = \mathop {\arg \min }\limits_\alpha  \frac{1}{2}\left\| {x - D\alpha } \right\|_2^2 + \frac{\lambda }{\mu }{\left\| \alpha  \right\|_0}.
\end{equation}

\cite{GSR} proved that Eq.(49) has an equivalent form as
\begin{equation}
\alpha  = \mathop {\min }\limits_{{\alpha _G}} \sum\limits_{k = 1}^M {\left( {\frac{1}{2}\left\| {{x_{{G_k}}} - {D_{{G_k}}}{\alpha _{{G_k}}}} \right\|_2^2 + \tau {{\left\| {{\alpha _{{G_k}}}} \right\|}_0}} \right)},
\end{equation}
where $\tau  = \frac{{\lambda Q}}{{\mu N}}$ and $Q = n \times m \times M$. $M$ is the number of groups. Eq.(50) reveals the relationship between the regularization parameter $\tau $ and other parameters. Following this theorem, we assign ${\lambda _1}  = \frac{{\lambda Q}}{{\mu N}}$ in Eq.(28).

Considering that each image patch has roughly the same probability of appearing in a patch group, Eq.(48) can be solved by solving every $\alpha _G^i$ via Eq.(28) \cite{LIIC}. The summary of the proposed method is given as \textbf{Algorithm 1}.

\begin{algorithm}[!h]
	\caption{Joint Group and Residual Sparse Coding for CS (JGRSC-CS)}
	\begin{algorithmic}[1]
		\REQUIRE
		$y$:measurement;
		$\Phi $:measurement matrix;
		
		\textbf{Initialization}:
		\quad
		
		(1) Estimate an initial image ${x_{init}}$;
		
		(2) Set parameters $m$, $m$, $K$, ${\sigma _n}$, $b$, $\lambda $, $\mu$;
		
		\FOR{$i = 1, \ldots ,Max\_Iter$}
		\STATE Compute $x$ via Eq.(43);
		\FOR{$j = 1, \ldots M$}
		\STATE (1) Group ${x_G}$ for each image patch;
		\STATE (2) Compute ${x_{Gr}}$;
		\STATE (3) Select the external dictionary via Eq.(30);
		\STATE (4) Compute ${\alpha _{Gr}}$ via Eq.(34);
		\STATE (5) Compute ${\alpha _G}$ via Eq.(38);
		\ENDFOR
		
		\STATE Update $b$ via Eq.(43)
				
		\ENDFOR       
		
		\ENSURE
		The reconstructed image ${x_{Re}}$
		
	\end{algorithmic}
	
\end{algorithm}

\section{Experimental results and analysis}

In this section, we present the performance of the proposed method. The measurement matrix is obtained by generating a Gaussian random matrix of size $32 \times 32$. In the training stage, the external dictionary is trained from the Kodak PhotoCD Dataset\footnote{http://r0k.us/graphics/kodak/}, and the number of Gaussian components $K$ is 64. In the recovery stage, the number of similar patches is set to $60$, and the size of patch $\sqrt n $, $\lambda $, $\mu $ are set to $(6,\ 0.082,\ 0.0025)$, $(8,\ 0.146,\ 0.0025)$, $(8,\ 0.146,\ 0.0025)$ when subrate=0.1, 0.2, 0.3, respectively. The maximum iteration is $120$, and the algorithm will terminate until the maximum iteration number is reached or the PSNR begins to decrease.

\subsection{Comparison with other methods}
\begin{figure}[H]
	\centering
	\subfigure[]{\includegraphics[width=0.24\textwidth]{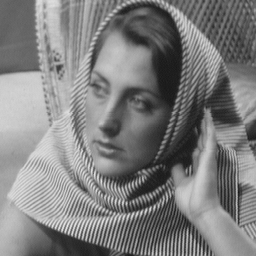}}
	\subfigure[]{\includegraphics[width=0.24\textwidth]{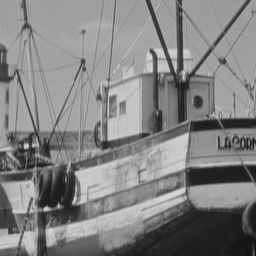}}
	\subfigure[]{\includegraphics[width=0.24\textwidth]{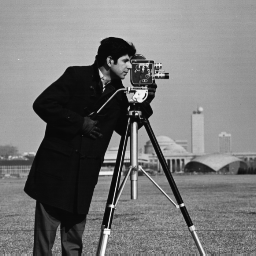}}
	\subfigure[]{\includegraphics[width=0.24\textwidth]{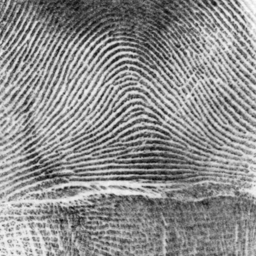}}
	\subfigure[]{\includegraphics[width=0.24\textwidth]{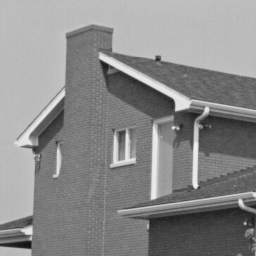}}
	\subfigure[]{\includegraphics[width=0.24\textwidth]{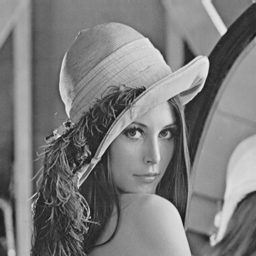}}
	\subfigure[]{\includegraphics[width=0.24\textwidth]{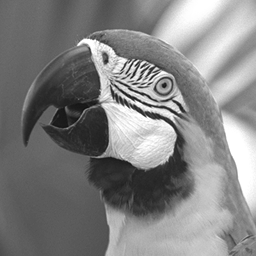}}
	\caption{Seven test images. (a) Barbara. (b) Boats. (c) Cameraman. (d) Fingerprint. (e) House. (f) Lena. (g) Parrots. }
	\label{testimage}
\end{figure}

We compare our method with several representative methods: BCS\cite{BCS}, MH-BCS\cite{MH}, RCoS\cite{CoS}, SGSR\cite{SGSR}, ALSB\cite{ALSB}, GSR-NCR\cite{GSR-NCR}. BCS and MH-BCS are block compressive sensing methods with fixed bases; RCoS combines 2D sparsity with 3D sparsity; ALSB is a patch-based method; SGSR and GSR-NCR are group-based methods, and the difference between them is that SGSR uses the ${l_0}$ norm to constrain the sparse coefficient, while GSR-NCR utilizes the non-convex ${l_p}$ norm. Seven test images are shown in Fig. \ref{testimage}, and PSNR as well as FSIM \cite{FSIM} are calculated to evaluate the quality of reconstructed images.

The PSNR and FSIM results are shown in Table \ref{table1}-\ref{table2}, and the highest score is marked in bold. From the tables, we can see that the proposed method achieves the highest PSNR and FSIM in most cases. Specifically, the average PSNR gain of the proposed JGRSC-CS method over BCS, MH, RCoS, SGSR, ALSB, GSR-NCR are 5.22dB, 2.28dB, 2.62dB, 0.80dB, 0.86dB and 0.34dB, respectively. The average improvements for FSIM over other method is 0.0885, 0.0295, 0.0514, 0.0102, 0.0124 and 0.0067.

\begin{figure}[H]
	\centering
	\subfigure[]{\includegraphics[width=0.24\textwidth]{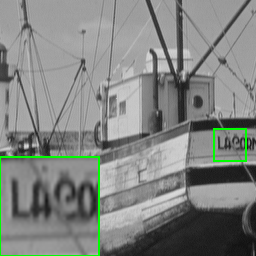}}
	\subfigure[]{\includegraphics[width=0.24\textwidth]{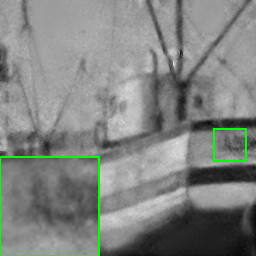}}
	\subfigure[]{\includegraphics[width=0.24\textwidth]{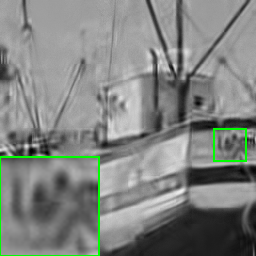}}
	\subfigure[]{\includegraphics[width=0.24\textwidth]{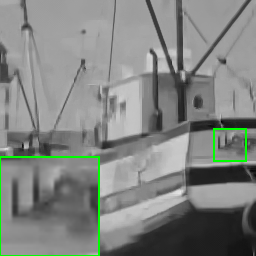}}
	\subfigure[]{\includegraphics[width=0.24\textwidth]{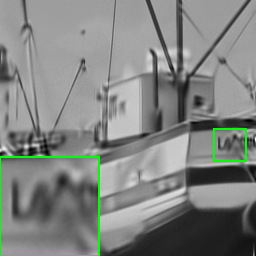}}
	\subfigure[]{\includegraphics[width=0.24\textwidth]{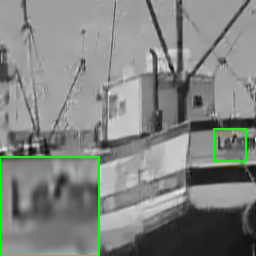}}
	\subfigure[]{\includegraphics[width=0.24\textwidth]{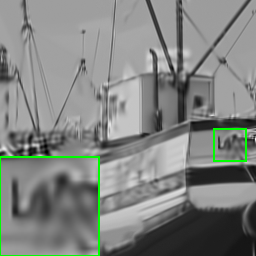}}
	\subfigure[]{\includegraphics[width=0.24\textwidth]{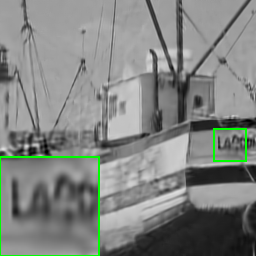}}
	\caption{Reconstruction of Boats with subrate=0.1. (a) Original image (b) BCS (PSNR=24.52dB, FSIM=0.8029); (c) MH (PSNR=26.11dB, FSIM=0.8489); (d) RCoS (PSNR=27.85dB, FSIM=0.8765); (e) SGSR (PSNR=27.74dB, FSIM=0.8918); (f) ALSB (PSNR=28.12dB, FSIM=0.8934); (g) GSR-NCR(PSNR=27.96dB, FSIM=0.8980); (h) the proposed JGRSC-CS (PSNR=\textbf{28.44dB}, FSIM=\textbf{0.9049}).}
	\label{boats}
\end{figure}

\begin{figure}[H]
	\centering
	\subfigure[]{\includegraphics[width=0.24\textwidth]{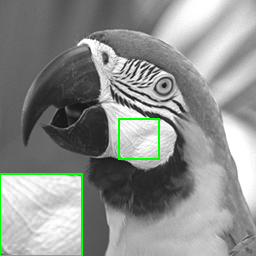}}
	\subfigure[]{\includegraphics[width=0.24\textwidth]{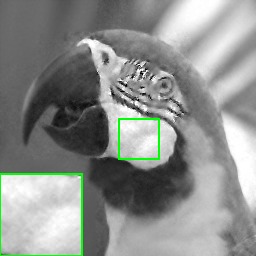}}
	\subfigure[]{\includegraphics[width=0.24\textwidth]{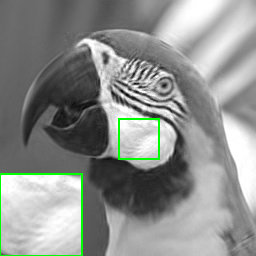}}
	\subfigure[]{\includegraphics[width=0.24\textwidth]{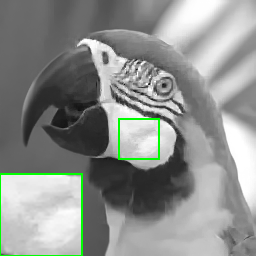}}
	\subfigure[]{\includegraphics[width=0.24\textwidth]{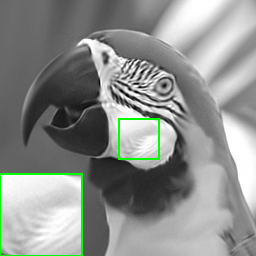}}
	\subfigure[]{\includegraphics[width=0.24\textwidth]{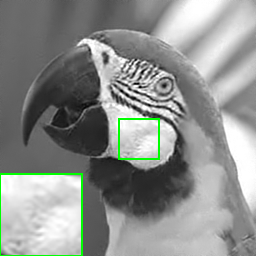}}
	\subfigure[]{\includegraphics[width=0.24\textwidth]{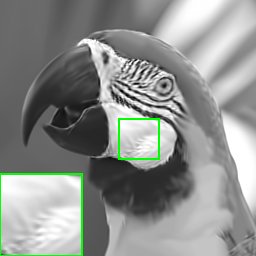}}
	\subfigure[]{\includegraphics[width=0.24\textwidth]{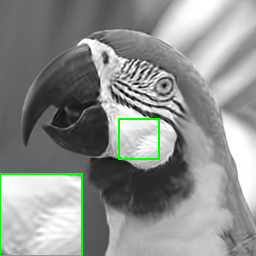}}
	\caption{Reconstruction of Parrots with subrate=0.2. (a) Original image (b) BCS (PSNR=26.29dB, FSIM=0.9188); (c) MH (PSNR=29.23dB, FSIM=0.9405); (d) RCoS (PSNR=28.61dB, FSIM=0.9311); (e) SGSR (PSNR=30.55dB, FSIM=0.9457); (f) ALSB (PSNR=29.73dB, FSIM=0.9460); (g) GSR-NCR(PSNR=30.18dB, FSIM=0.9435); (h) the proposed JGRSC-CS (PSNR=\textbf{30.82dB}, FSIM=\textbf{0.9539}).}
	\label{parrots}
\end{figure}

\begin{figure}[H]
	\centering
	\subfigure[]{\includegraphics[width=0.24\textwidth]{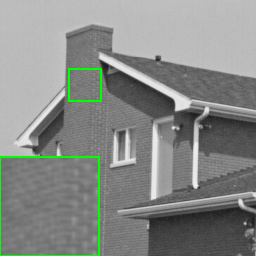}}
	\subfigure[]{\includegraphics[width=0.24\textwidth]{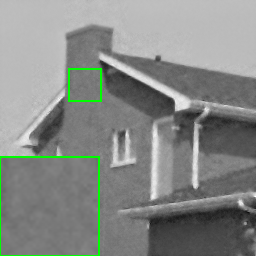}}
	\subfigure[]{\includegraphics[width=0.24\textwidth]{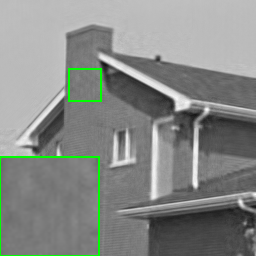}}
	\subfigure[]{\includegraphics[width=0.24\textwidth]{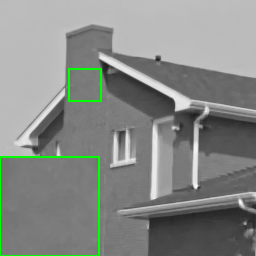}}
	\subfigure[]{\includegraphics[width=0.24\textwidth]{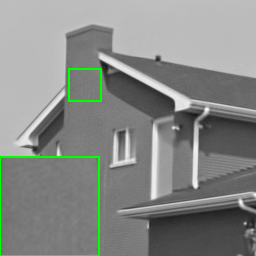}}
	\subfigure[]{\includegraphics[width=0.24\textwidth]{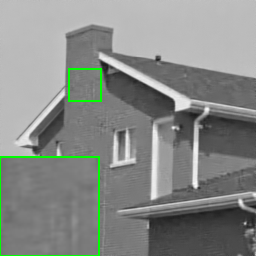}}
	\subfigure[]{\includegraphics[width=0.24\textwidth]{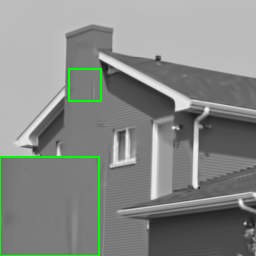}}
	\subfigure[]{\includegraphics[width=0.24\textwidth]{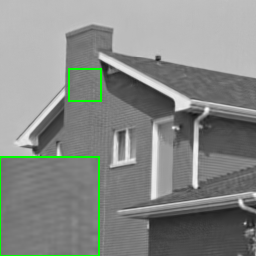}}
	\caption{Reconstruction of House with subrate=0.2. (a) Original image (b) BCS (PSNR=30.58dB, FSIM=0.9014); (c) MH (PSNR=33.84dB, FSIM=0.9389); (d) RCoS (PSNR=35.22dB, FSIM=0.9388); (e) SGSR (PSNR=35.81dB, FSIM=0.9502); (f) ALSB (PSNR=35.86dB, FSIM=0.9540); (g) GSR-NCR(PSNR=36.56dB, FSIM=0.9507); (h) the proposed JGRSC-CS (PSNR=\textbf{37.18dB}, FSIM=\textbf{0.9670}).}
	\label{house}
\end{figure}

\begin{figure}[H]
	\centering
	\subfigure[]{\includegraphics[width=0.24\textwidth]{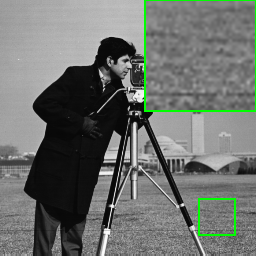}}
	\subfigure[]{\includegraphics[width=0.24\textwidth]{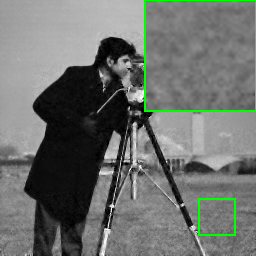}}
	\subfigure[]{\includegraphics[width=0.24\textwidth]{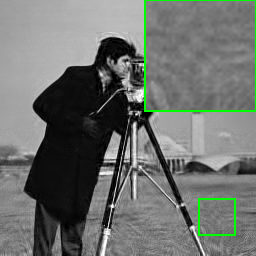}}
	\subfigure[]{\includegraphics[width=0.24\textwidth]{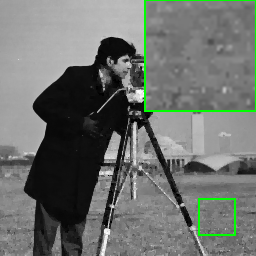}}
	\subfigure[]{\includegraphics[width=0.24\textwidth]{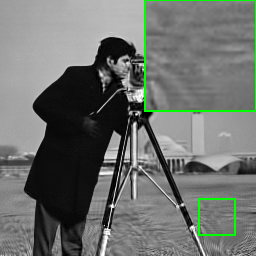}}
	\subfigure[]{\includegraphics[width=0.24\textwidth]{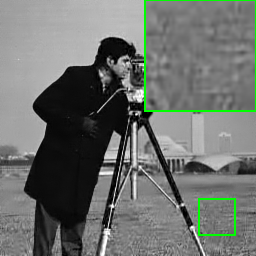}}
	\subfigure[]{\includegraphics[width=0.24\textwidth]{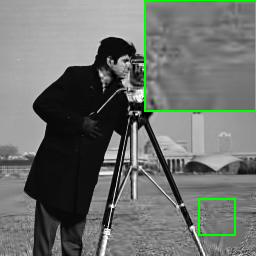}}
	\subfigure[]{\includegraphics[width=0.24\textwidth]{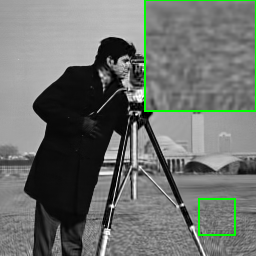}}
	\caption{Reconstruction of Cameraman with subrate=0.3. (a) Original image (b) BCS (PSNR=27.12dB, FSIM=0.8798); (c) MH (PSNR=28.08dB, FSIM=0.8938); (d) RCoS (PSNR=27.98dB, FSIM=0.9089); (e) SGSR (PSNR=28.89dB, FSIM=0.9219); (f) ALSB (PSNR=29.01dB, FSIM=0.9190); (g) GSR-NCR(PSNR=29.37dB, FSIM=0.9305); (h) the proposed JGRSC-CS (PSNR=\textbf{29.54dB}, FSIM=\textbf{0.9358}).}
	\label{cameraman}
\end{figure}

\begin{figure}[H]
	\centering
	\subfigure[]{\includegraphics[width=0.24\textwidth]{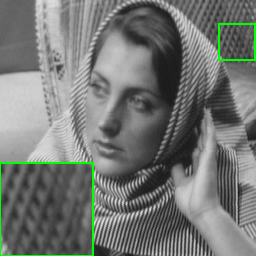}}
	\subfigure[]{\includegraphics[width=0.24\textwidth]{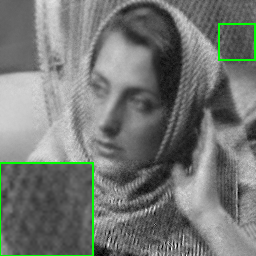}}
	\subfigure[]{\includegraphics[width=0.24\textwidth]{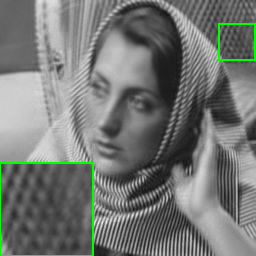}}
	\subfigure[]{\includegraphics[width=0.24\textwidth]{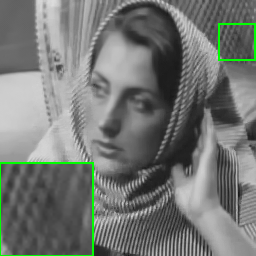}}
	\subfigure[]{\includegraphics[width=0.24\textwidth]{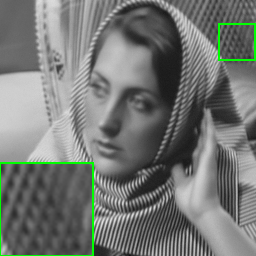}}
	\subfigure[]{\includegraphics[width=0.24\textwidth]{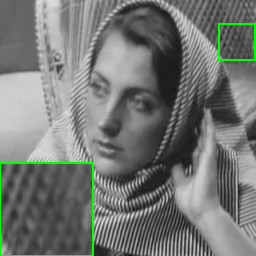}}
	\subfigure[]{\includegraphics[width=0.24\textwidth]{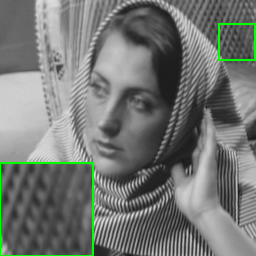}}
	\subfigure[]{\includegraphics[width=0.24\textwidth]{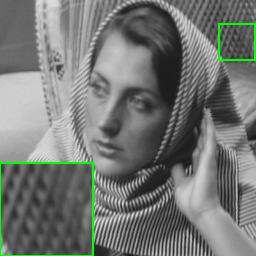}}
	\caption{Reconstruction of Barbara with subrate=0.3. (a) Original image (b) BCS (PSNR=25.70dB, FSIM=0.8780); (c) MH (PSNR=33.00dB, FSIM=0.9588); (d) RCoS (PSNR=30.06dB, FSIM=0.9398); (e) SGSR (PSNR=35.91dB, FSIM=0.9762); (f) ALSB (PSNR=34.76dB, FSIM=0.9718); (g) GSR-NCR(PSNR=\textbf{37.19dB}, FSIM=\textbf{0.9816}); (h) the proposed JGRSC-CS (PSNR=37.14dB, FSIM=\textbf{0.9816}).}
	\label{barbara}
\end{figure}

Some results are shown in Fig. \ref{boats}- \ref{barbara}. It is evident that the proposed method outperforms other methods in terms of visual quality. For instance, the text on the boats in Fig. \ref{boats}, the texture on the wall in Fig. \ref{house} and the ripples on the water in Fig. \ref{cameraman} are reconstructed sharply, while other methods still suffer from over-smooth or artifacts.

\begin{table}[H]
	\caption{The PSNR (dB) results of various methods}
	\resizebox{\textwidth}{!}{
		\begin{tabular}{|c|c|c|c|c|c|c|c|c|c|}
			\hline
			Subrate              & Method   & House          & Barbara        & Boats          & C.man      & Lena           & Parrots        & F.print    & Average        \\ \hline
			\multirow{7}{*}{0.1} & BCS      & 26.90          & 22.80          & 24.52          & 21.60          & 25.20          & 23.45          & 17.15          & 23.09          \\ \cline{2-10} 
			& MH       & 30.28          & 26.73          & 26.11          & 22.13          & 26.13          & 25.34          & 20.08          & 25.26          \\ \cline{2-10} 
			& RCoS     & 32.06          & 23.78          & 27.85          & 22.97          & 27.53          & 25.60          & 16.30          & 25.16          \\ \cline{2-10} 
			& SGSR     & 32.77          & \textbf{28.70} & 27.74          & 22.60          & 27.10          & 26.03          & 20.50          & 26.49          \\ \cline{2-10} 
			& ALSB     & 32.38          & 27.30          & 28.12          & 22.97          & 27.04          & 26.03          & 20.68          & 26.36          \\ \cline{2-10} 
			& GSR-NCR  & \textbf{32.83} & 28.70          & 27.96          & 22.50          & 27.02          & 26.03          & 20.50          & 26.51          \\ \cline{2-10} 
			& proposed & 32.80          & 28.66          & \textbf{28.44} & \textbf{23.40} & \textbf{27.82} & \textbf{27.07} & \textbf{20.72} & \textbf{26.99} \\ \hline
			\multirow{7}{*}{0.2} & BCS      & 30.58          & 24.31          & 27.05          & 24.65          & 28.04          & 26.29          & 18.55          & 25.64          \\ \cline{2-10} 
			& MH       & 33.84          & 30.82          & 29.91          & 25.88          & 29.81          & 29.23          & 23.17          & 28.95          \\ \cline{2-10} 
			& RCoS     & 35.22          & 27.19          & 31.42          & 25.68          & 30.36          & 28.61          & 19.64          & 28.30          \\ \cline{2-10} 
			& SGSR     & 35.81          & 33.45          & 32.41          & 26.53          & 30.89          & 30.55          & 23.62          & 30.47          \\ \cline{2-10} 
			& ALSB     & 35.86          & 31.98          & 33.27          & 26.65          & 30.73          & 29.73          & 23.64          & 30.27          \\ \cline{2-10} 
			& GSR-NCR  & 36.56          & 33.92          & 33.30          & 26.30          & 30.87          & 30.18          & 23.67          & 30.69          \\ \cline{2-10} 
			& proposed & \textbf{37.18} & \textbf{34.48} & \textbf{33.49} & \textbf{27.00} & \textbf{31.27} & \textbf{30.82} & \textbf{23.91} & \textbf{31.16} \\ \hline
			\multirow{7}{*}{0.3} & BCS      & 32.87          & 25.70          & 28.93          & 27.12          & 30.08          & 28.62          & 20.05          & 27.62          \\ \cline{2-10} 
			& MH       & 35.69          & 33.00          & 32.25          & 28.08          & 31.99          & 31.01          & 24.73          & 30.96          \\ \cline{2-10} 
			& RCoS     & 36.87          & 30.06          & 34.32          & 27.98          & 32.41          & 30.53          & 22.74          & 30.70          \\ \cline{2-10} 
			& SGSR     & 37.37          & 35.91          & 35.22          & 28.89          & 33.26          & 32.16          & 25.84          & 32.66          \\ \cline{2-10} 
			& ALSB     & 38.25          & 34.76          & 36.59          & 29.01          & 33.30          & 31.98          & 25.81          & 32.81          \\ \cline{2-10} 
			& GSR-NCR  & 39.38          & \textbf{37.19} & \textbf{37.27} & 29.37          & 33.94          & 33.07          & \textbf{26.35} & 33.80          \\ \cline{2-10} 
			& proposed & \textbf{39.45} & 37.14          & 36.94          & \textbf{29.54} & \textbf{33.97} & \textbf{33.73} & 26.31          & \textbf{33.87} \\ \hline
		\end{tabular}
	}
\label{table1}
\end{table}

\begin{table}[H]
	\caption{The FSIM results of various methods}
	\resizebox{\textwidth}{!}{
		\begin{tabular}{|c|c|c|c|c|c|c|c|c|c|}
			\hline
			Subrate              & Method   & House           & Barbara         & Boats           & C.man       & Lena            & Parrots         & F.print     & Average         \\ \hline
			\multirow{7}{*}{0.1} & BCS      & 0.8455          & 0.7891          & 0.8029          & 0.7605          & 0.8553          & 0.8786          & 0.6165          & 0.7926          \\ \cline{2-10} 
			& MH       & 0.8935          & 0.8909          & 0.8489          & 0.7692          & 0.8913          & 0.8981          & 0.8512          & 0.8633          \\ \cline{2-10} 
			& RCoS     & 0.8989          & 0.8065          & 0.8765          & 0.7942          & 0.8863          & 0.8919          & 0.6027          & 0.8224          \\ \cline{2-10} 
			& SGSR     & 0.9187          & 0.9149          & 0.8918          & 0.8065          & 0.9061          & 0.9142          & 0.8672          & 0.8885          \\ \cline{2-10} 
			& ALSB     & 0.9121          & 0.8945          & 0.8934          & 0.8021          & 0.8965          & 0.9105          & 0.8682          & 0.8825          \\ \cline{2-10} 
			& GSR-NCR  & 0.9132          & \textbf{0.9215} & 0.8980          & 0.8012          & 0.9106          & 0.919           & \textbf{0.8688} & 0.8903          \\ \cline{2-10} 
			& proposed & \textbf{0.9272} & 0.9207          & \textbf{0.9049} & \textbf{0.8335} & \textbf{0.9166} & \textbf{0.9279} & 0.8649          & \textbf{0.8994} \\ \hline
			\multirow{7}{*}{0.2} & BCS      & 0.9014          & 0.8429          & 0.8640          & 0.8357          & 0.9053          & 0.9188          & 0.7378          & 0.8580          \\ \cline{2-10} 
			& MH       & 0.9389          & 0.9394          & 0.9159          & 0.8552          & 0.9348          & 0.9405          & 0.9103          & 0.9193          \\ \cline{2-10} 
			& RCoS     & 0.9388          & 0.8977          & 0.9348          & 0.8645          & 0.9331          & 0.9311          & 0.7923          & 0.8989          \\ \cline{2-10} 
			& SGSR     & 0.9502          & 0.9615          & 0.9465          & 0.8847          & 0.9472          & 0.9457          & 0.9207          & 0.9366          \\ \cline{2-10} 
			& ALSB     & 0.9540          & 0.9502          & 0.9522          & 0.8759          & 0.9440          & 0.9460          & 0.9208          & 0.9347          \\ \cline{2-10} 
			& GSR-NCR  & 0.9507          & 0.9643          & 0.9526          & 0.8797          & 0.9470          & 0.9435          & 0.9225          & 0.9372          \\ \cline{2-10} 
			& proposed & \textbf{0.9670} & \textbf{0.9692} & \textbf{0.9569} & \textbf{0.9003} & \textbf{0.9546} & \textbf{0.9539} & \textbf{0.9272} & \textbf{0.9470} \\ \hline
			\multirow{7}{*}{0.3} & BCS      & 0.9298          & 0.8780          & 0.8995          & 0.8798          & 0.9327          & 0.9418          & 0.8191          & 0.8972          \\ \cline{2-10} 
			& MH       & 0.9569          & 0.9588          & 0.9439          & 0.8938          & 0.9538          & 0.9563          & 0.9331          & 0.9424          \\ \cline{2-10} 
			& RCoS     & 0.9560          & 0.9398          & 0.9615          & 0.9089          & 0.9555          & 0.9501          & 0.8937          & 0.9379          \\ \cline{2-10} 
			& SGSR     & 0.9648          & 0.9762          & 0.9684          & 0.9219          & 0.9643          & 0.9594          & 0.9480          & 0.9576          \\ \cline{2-10} 
			& ALSB     & 0.9727          & 0.9718          & 0.9748          & 0.9190          & 0.9650          & 0.9620          & 0.9471          & 0.9589          \\ \cline{2-10} 
			& GSR-NCR  & 0.9795          & 0.9816          & \textbf{0.9783} & 0.9305          & 0.9715          & 0.9660          & \textbf{0.9534} & 0.9658          \\ \cline{2-10} 
			& proposed & \textbf{0.9795} & \textbf{0.9816} & 0.9773          & \textbf{0.9358} & \textbf{0.9715} & \textbf{0.9693} & 0.9530          & \textbf{0.9669} \\ \hline
	\end{tabular}}
\label{table2}
\end{table}

\subsection{Convergence Analysis}
\begin{figure}[H]
	\centering
	\subfigure[]{\includegraphics[width=0.45\textwidth]{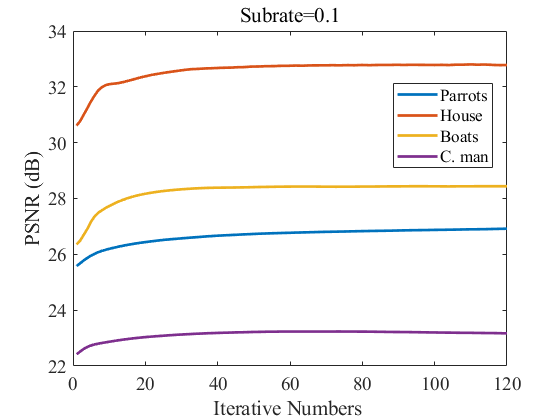}}
	\subfigure[]{\includegraphics[width=0.45\textwidth]{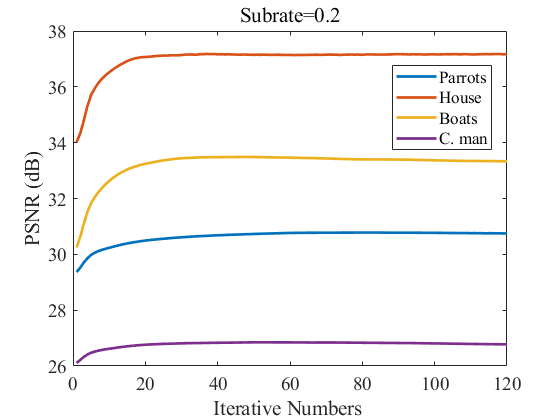}}

	\caption{Evolutions of PSNR versus iteration number for four test images. (a) Subrate=0.1; (b) Subrate=0.2. }
	\label{convergence}
\end{figure}

Fig. \ref{convergence} shows the PSNR curves of four test images with subrate= 0.1 and 0.2. It is obvious that with the iteration number increases, all the curves increase rapidly, and then gradually become stable. This also proves the robustness and effectiveness of the proposed method.

\subsection{Computational complexity}
All experiments are implemented under Matlab R2018b environment on a machine with Intel Core i5-6500 CPU of 3.2Ghz and 8.0 GB RAM. We calculate the average run time of reconstructing an image in the case of subrate=0.2, and the results are shown in Table \ref{time}. We can see that BCS is the fastest method, while its performance is the worst. JGRSC has comparable time consumption with SGSR and ALSB. This is because the proposed method use the SVD decomposition in every iteration, which has high computational complexity. However, this can be accelerated by parallel computing.

\begin{table}[H]
	\caption{Average run time (seconds) with subrate=0.2}
	\begin{tabular}{|c|c|c|c|c|c|c|c|}
		\hline
		Method & BCS  & MH    & RCoS    & SGSR   & ALSB   & GSR-NCR & Proposed \\ \hline
		Time   & 5.42 & 25.76 & 2521.22 & 491.56 & 573.24 & 2677.69 & 576.43   \\ \hline
	\end{tabular}
\label{time}
\end{table}

\section{Conclusion}
In this paper, we proposed a joint group and residual sparse coding method for image compressive sensing (JGRSC-CS). For a patch group, its residual is coded using an external dictionary that learned from clean images, and the whole group is coded with adaptively SVD dictionary. An effective framework is also present to solve the optimization problem. Experimental results show that the proposed JGRSC-CS not only outperforms many existing methods in terms of PSNR and FSIM, but also has better visual quality.

\section*{References}

\bibliography{JGRSC}

\begin{thebibliography}{10}
\expandafter\ifx\csname url\endcsname\relax
  \def\url#1{\texttt{#1}}\fi
\expandafter\ifx\csname urlprefix\endcsname\relax\def\urlprefix{URL }\fi
\expandafter\ifx\csname href\endcsname\relax
  \def\href#1#2{#2} \def\path#1{#1}\fi

\bibitem{D.L.Donoho2006}
D.~L. Donoho, Compressed sensing, IEEE Transactions on Information Theory
  52~(4) (2006) 1289--1306.
\newblock \href {http://dx.doi.org/10.1109/TIT.2006.871582}
  {\path{doi:10.1109/TIT.2006.871582}}.

\bibitem{E.J.Candes}
E.~J. Candes, J.~Romberg, T.~Tao, Robust uncertainty principles: exact signal
  reconstruction from highly incomplete frequency information, IEEE
  Transactions on Information Theory 52~(2) (2006) 489--509.
\newblock \href {http://dx.doi.org/10.1109/TIT.2005.862083}
  {\path{doi:10.1109/TIT.2005.862083}}.

\bibitem{Qaisar2013}
S.~Qaisar, R.~M. Bilal, W.~Iqbal, M.~Naureen, S.~Lee, Compressed sensing: From
  theory to applications, a survey, Journal fo Communications and Networks
  15~(5) (2013) 443--456.
\newblock \href {http://dx.doi.org/10.1109/JCN.2013.000083}
  {\path{doi:10.1109/JCN.2013.000083}}.

\bibitem{Singlepixelcam}
M.~F. Duarte, M.~A. Davenport, D.~Takhar, J.~N.~L. abd T.~Sun, K.~F. Kelly,
  R.~G. Baraniuk, Signal-pixel imaging via compressive sampling, IEEE Signal
  Processing Magazine 25~(2) (2008) 83--91.
\newblock \href {http://dx.doi.org/10.1109/MSP.2007.914730}
  {\path{doi:10.1109/MSP.2007.914730}}.

\bibitem{Bajwa2010}
W.~U. Bajwa, J.~Haupt, A.~M. Sayeed, R.~Nowak, Compressed channel sensing: A
  new approach to estimating sparse multipath channels, Proceedings of the IEEE
  98~(6) (2010) 1058--1076.
\newblock \href {http://dx.doi.org/10.1109/JPROC.2010.2042415}
  {\path{doi:10.1109/JPROC.2010.2042415}}.

\bibitem{Luo2009}
C.~Luo, F.~Wu, J.~Sun, C.~Chen, Compressive data gathering for large-scale
  wireless sensor networks, in: 2009 International Conference on Mobile
  Computing and Networking (MobiCom), ACM, 2009, pp. 145--156.
\newblock \href {http://dx.doi.org/10.1145/1614320.1614337}
  {\path{doi:10.1145/1614320.1614337}}.

\bibitem{Lustig2008}
M.~Lustig, D.~L. Donoho, J.~M. Santos, J.~M. Pauly, Compressed sensing {MRI},
  IEEE Signal Processing Magazine 25~(2) (2008) 72--82.
\newblock \href {http://dx.doi.org/10.1109/MSP.2007.914728}
  {\path{doi:10.1109/MSP.2007.914728}}.

\bibitem{Alonso2010}
M.~T. Alonso, P.~Lopez-Dekker, J.~J. Mallorqui, A novel strategy for radar
  imaging based on compressive sensing, IEEE Transactions on Geoscience and
  Remote Sensing 48~(12) (2010) 4285--4295.
\newblock \href {http://dx.doi.org/10.1109/TGRS.2010.2051231}
  {\path{doi:10.1109/TGRS.2010.2051231}}.

\bibitem{SBI}
T.~Goldstein, S.~Osher, The split bregman method for l1 regularized problems,
  SIAM Journal on Image Sciences 2~(2) (2009) 323--343.
\newblock \href {http://dx.doi.org/10.1137/080725891}
  {\path{doi:10.1137/080725891}}.

\bibitem{TVAL3}
C.~Li, W.~Yin, H.~Jiang, Y.~Zhang, An efficient augmented lagrangian method
  with applications to total variation minimization, Computational Optimization
  and Applications 56~(3) (2013) 507--530.
\newblock \href {http://dx.doi.org/10.1007/s10589-013-9576-1}
  {\path{doi:10.1007/s10589-013-9576-1}}.

\bibitem{reweightedl1}
E.~J. Candes, M.~B. Wakin, S.~P. Boyd, Enhancing sparsity by reweighted l1
  minimization, Journal of Fourier Analysis and Application 12~(5-6) (2008)
  877--905.
\newblock \href {http://dx.doi.org/10.1007/s00041-008-9045-x}
  {\path{doi:10.1007/s00041-008-9045-x}}.

\bibitem{ImprovedTV}
J.~Zhang, S.~Liu, R.~Xiong, S.~Ma, D.~Zhao, Improved total variation based
  image compressive sensing recovery by nonlocal regularization, in: 2013 IEEE
  International Symposium on Circuits and Systems (ISCAS), IEEE, 2013, pp.
  2836--2839.
\newblock \href {http://dx.doi.org/10.1109/ISCAS.2013.6572469}
  {\path{doi:10.1109/ISCAS.2013.6572469}}.

\bibitem{frtv}
G.~Chen, J.~Zhang, D.~Li, Fractional-order total variation combined with
  sparsifying transforms for compressive sensing sparse image reconstruction,
  Journal of Visual Communication and Image Representation 38 (2016) 407--422.
\newblock \href {http://dx.doi.org/10.1016/j.jvcir.2016.03.018}
  {\path{doi:10.1016/j.jvcir.2016.03.018}}.

\bibitem{BM3D}
K.~Dabov, A.~Foi, V.~Katkovnik, K.~Egiazarian, Image denoising by sparse 3-d
  transform-domain collaborative filtering, IEEE Transactions on Image
  Processing 16~(8) (2007) 2080--2095.
\newblock \href {http://dx.doi.org/10.1109/TIP.2007.901238}
  {\path{doi:10.1109/TIP.2007.901238}}.

\bibitem{BM3DCS}
K.~Egiazarian, A.~Foi, V.~Katkovnik, Compressed sensing image reconstruction
  via recursive spatially adaptive filtering, in: 2007 International Conference
  on Image Processing (ICIP), IEEE, 2007, pp. I--549--I--552.
\newblock \href {http://dx.doi.org/10.1109/ICIP.2007.4379013}
  {\path{doi:10.1109/ICIP.2007.4379013}}.

\bibitem{Mairal2009}
J.~Mairal, F.~Bach, J.~Ponce, G.~Sapiro, A.~Zisserman, Non-local sparse models
  for image restoration, in: 2009 IEEE International Conference on Computer
  Vision (ICCV), IEEE, 2009, pp. 2272--2279.
\newblock \href {http://dx.doi.org/10.1109/ICCV.2009.5459452}
  {\path{doi:10.1109/ICCV.2009.5459452}}.

\bibitem{CoS}
J.~Zhang, D.~Zhao, C.~Zhao, R.~Xiong, S.~Ma, W.~Gao, Image compressive sensing
  recovery via collaborative sparsity, IEEE Journal on Emerging and Selected
  Topic in Circuits and Systems 2~(3) (2012) 380--391.
\newblock \href {http://dx.doi.org/10.1109/JETCAS.2012.2220391}
  {\path{doi:10.1109/JETCAS.2012.2220391}}.

\bibitem{JASRCS}
N.~Eslahi, A.~Aghagolzadeh, Compressive sensing image restoration using
  adaptive curvelet thresholding and nonlocal sparse regularization, IEEE
  Transactions on Image Processing 25~(7) (2016) 3126--3140.
\newblock \href {http://dx.doi.org/10.1109/TIP.2016.2562563}
  {\path{doi:10.1109/TIP.2016.2562563}}.

\bibitem{Elad2006}
M.~Elad, M.~Aharon, Image denoising via sparse and redundant representations
  over learned dictionaries, IEEE Transactions on Image Processing 15~(12)
  (2006) 3736--3745.
\newblock \href {http://dx.doi.org/10.1109/TIP.2006.881969}
  {\path{doi:10.1109/TIP.2006.881969}}.

\bibitem{Dong20121109}
W.~Dong, G.~Shi, X.~Li, L.~Zhang, X.~Wu, Image reconstruction with locally
  adaptive sparsity and nonlocal robust regularization, Signal Processing:
  Image Communication 27~(10) (2012) 1109--1122.
\newblock \href {http://dx.doi.org/10.1016/j.image.2012.09.003}
  {\path{doi:10.1016/j.image.2012.09.003}}.

\bibitem{NCSR}
W.~Dong, L.~Zhang, G.~Shi, X.~Li, Nonlocally centralized sparse representation
  for image restoration, IEEE Transactions on Image Processing 22~(4) (2013)
  1620--1630.
\newblock \href {http://dx.doi.org/10.1109/TIP.2012.2235847}
  {\path{doi:10.1109/TIP.2012.2235847}}.

\bibitem{Eslahi201688}
N.~Eslahi, A.~Aghagolzadeh, S.~M.~H. Andargoli, Image/video compressive sensing
  recovery using joint adaptive sparsity measure, Neurocomputing 200 (2016)
  88--109.
\newblock \href {http://dx.doi.org/10.1016/j.neucom.2014.05.088}
  {\path{doi:10.1016/j.neucom.2014.05.088}}.

\bibitem{ALSB}
J.~Zhang, C.~Zhao, D.~Zhao, W.~Gao, Image compressive sensing recovery using
  adaptively learned sparsifying basis via l0 minimization, Signal Processing
  103 (2014) 114--126.
\newblock \href {http://dx.doi.org/10.1016/j.sigpro.2013.09.025}
  {\path{doi:10.1016/j.sigpro.2013.09.025}}.

\bibitem{NIPS2010_3997}
P.~Garrigues, B.~A. Olshausen, Group sparse coding with laplacian scale mixture
  prior, in: 2010 Advances in Neural Information Processing Systems (NIPS),
  Curran Associates, Inc., 2010, pp. 676--684.
\newblock \href {http://dx.doi.org/10.1007/978-3-319-59463-7_51}
  {\path{doi:10.1007/978-3-319-59463-7_51}}.

\bibitem{SGSR}
J.~Zhang, D.~Zhao, F.~Jiang, W.~Gao, Structural group sparse representation for
  image compressive sensing recovery, in: 2013 Data Compression Conference
  (DCC), IEEE, 2013, pp. 331--340.
\newblock \href {http://dx.doi.org/10.1109/DCC.2013.41}
  {\path{doi:10.1109/DCC.2013.41}}.

\bibitem{GSR}
J.~Zhang, D.~Zhao, W.~Gao, Group-based sparse representation for image
  restoration, IEEE Transactions on Image Processing 23~(8) (2014) 3336--3351.
\newblock \href {http://dx.doi.org/10.1109/TIP.2014.2323127}
  {\path{doi:10.1109/TIP.2014.2323127}}.

\bibitem{GSR-NCR}
Z.~Zha, X.~Zhang, Q.~Wang, L.~Tang, X.~Liu, Group-based sparse representation
  for image compressive sensing reconstruction with non-convex regularization,
  Neurocomputing 296 (2018) 55--63.
\newblock \href {http://dx.doi.org/10.1016/j.neucom.2018.03.027}
  {\path{doi:10.1016/j.neucom.2018.03.027}}.

\bibitem{Zoran2011GMM}
D.~Zoran, Y.~Weiss, From learning models of natural image patches to whole
  image restoration, in: 2011 International Conference on Computer Vision
  (ICCV), IEEE, 011, pp. 479--486.
\newblock \href {http://dx.doi.org/10.1109/ICCV.2011.6126278}
  {\path{doi:10.1109/ICCV.2011.6126278}}.

\bibitem{Yu2012}
G.~Yu, G.~Sapiro, S.~Mallat, Solving inverse problems with piecewise linear
  estimators: From gaussian mixture models to structured sparsity, IEEE
  Transactions on Image Processing 21~(5) (2012) 2481--2499.
\newblock \href {http://dx.doi.org/10.1109/TIP.2011.2176743}
  {\path{doi:10.1109/TIP.2011.2176743}}.

\bibitem{yang2014compressive}
J.~Yang, X.~Liao, M.~Chen, L.~Carin, Compressive sensing of signals from a gmm
  with sparse precision matrices, in: 2014 Advances in Neural Information
  Processing Systems, Curran Associates, Inc., 2014, pp. 3194--3202.

\bibitem{yang2014video}
J.~Yang, X.~Yuan, X.~Liao, P.~Llull, D.~J. Brady, G.~Sapiro, L.~Carin, Video
  compressive sensing using gaussian mixture models, IEEE Transactions on Image
  Processing 23~(11) (2014) 4863--4878.
\newblock \href {http://dx.doi.org/10.1109/TIP.2014.2344294}
  {\path{doi:10.1109/TIP.2014.2344294}}.

\bibitem{GMMImage}
J.~Yang, X.Liao, X.~Yuan, P.~Llull, D.~J. Brady, G.~Sapiro, L.Carin,
  Compressive sensing by learning a gaussian mixture model from measurements,
  IEEE Transactions on Image Processing 24~(1) (2015) 106--109.
\newblock \href {http://dx.doi.org/10.1109/TIP.2014.2365720}
  {\path{doi:10.1109/TIP.2014.2365720}}.

\bibitem{PGD}
J.~Xu, L.~Zhang, W.~Zuo, D.~Zhang, X.~Feng, Patch group based nonlocal
  self-similarity prior learning for image denoising, in: 2015 IEEE
  International Conference on Computer Vision (ICCV), IEEE, 2015, pp. 244--252.
\newblock \href {http://dx.doi.org/10.1109/ICCV.2015.36}
  {\path{doi:10.1109/ICCV.2015.36}}.

\bibitem{xu2018}
J.~Xu, L.~Zhang, D.~Zhang, External prior guided internal prior learning for
  real-world noisy image denoising, IEEE Transactions on Image Processing
  27~(6) (2018) 2996--3010.
\newblock \href {http://dx.doi.org/10.1109/TIP.2018.2811546}
  {\path{doi:10.1109/TIP.2018.2811546}}.

\bibitem{Hard}
M.~Elad, M.~A.~T. Figueiredo, Y.~Ma, On the role of sparse and redundant
  representations in image processing, Proceedings of the IEEE 98~(6) (2010)
  972--982.
\newblock \href {http://dx.doi.org/10.1109/JPROC.2009.2037655}
  {\path{doi:10.1109/JPROC.2009.2037655}}.

\bibitem{LIIC}
H.~Liu, R.~Xiong, D.~Liu, S.~Ma, F.~Wu, W.~Gao, Image denoising via low rank
  regularization exploiting intra and inter patch correlation, IEEE
  Transactions on Circuits and Systems for Video Technology 28~(12) (2018)
  3321--3332.
\newblock \href {http://dx.doi.org/10.1109/TCSVT.2017.2759187}
  {\path{doi:10.1109/TCSVT.2017.2759187}}.

\bibitem{BCS}
S.~Mun, J.~E. Fowler, Block compressed sensing of images using directional
  transforms, in: 2009 International Conference on Image Processing (ICIP),
  IEEE, 2009, pp. 3021--3024.
\newblock \href {http://dx.doi.org/10.1109/ICIP.2009.5414429}
  {\path{doi:10.1109/ICIP.2009.5414429}}.

\bibitem{MH}
C.~Chen, E.~W. Tramel, J.~E. Fowler, Compressed-sensing recovery of images and
  video using multihypothesis predictions, in: 2011 Asilomar Conference on
  Signal, Systems and Computer (ASILOMAR), IEEE, 2011, pp. 1193--1198.
\newblock \href {http://dx.doi.org/10.1109/ACSSC.2011.6190204}
  {\path{doi:10.1109/ACSSC.2011.6190204}}.

\bibitem{FSIM}
L.~Zhang, L.~Zhang, X.~Mou, D.~Zhang, Fsim: A feature similarity index for
  image quality assessment, IEEE Transactions on Image Processing 20~(8) (2011)
  2378--2386.
\newblock \href {http://dx.doi.org/10.1109/TIP.2011.2109730}
  {\path{doi:10.1109/TIP.2011.2109730}}.

\end{thebibliography}

\end{document}